\documentclass[11pt]{article}

\usepackage[preprint]{acl}
\usepackage{enumitem}
\usepackage{times}
\usepackage{latexsym}
\usepackage{makecell}
\usepackage[T1]{fontenc}

\usepackage[utf8]{inputenc}

\usepackage{microtype}

\usepackage{inconsolata}

\usepackage{graphicx}
\usepackage{amsfonts}

\usepackage[english]{babel}
\usepackage{booktabs}
\usepackage{multirow}
\usepackage{graphicx}
\usepackage{amsmath}
\usepackage[section]{placeins}
\usepackage{adjustbox}
\usepackage{colortbl}  
\usepackage{xcolor}    
\definecolor{highlightrow}{gray}{0.92} 

\usepackage{booktabs}
\usepackage{array}
\usepackage{xurl}

\newcolumntype{L}[1]{>{\raggedright\arraybackslash}p{#1}}
\newcommand{\fsep}{\hspace{0pt}\textbar\hspace{0pt}}

\usepackage{tabularx}
\usepackage{array}
\newcolumntype{Y}{>{\raggedright\arraybackslash}X}


\title{Let the Agent Search: Autonomous Exploration Beats Rigid Workflows in Temporal Question Answering}

\author{
  \textbf{Xufei Lv}\textsuperscript{2,*}, 
  \textbf{Jiahui Yang}\textsuperscript{1,*}, 
  \textbf{Haoyuan Sun}\textsuperscript{2},
  \textbf{Xialin Su}\textsuperscript{1},
  \textbf{Zhiliang Tian}\textsuperscript{1}, 
  \textbf{Yifu Gao}\textsuperscript{1,$\dagger$}, 
  \textbf{Linbo Qiao}\textsuperscript{1,$\dagger$},
  \textbf{Houde Liu}\textsuperscript{2,$\dagger$}
\\
  \textsuperscript{1}National University of Defense Technology 
  \textsuperscript{2}Tsinghua University \\
  \small{\texttt{lvxf24@mails.tsinghua.edu.cn, \{yangjiahui, gaoyifu, qiao.linbo\}@nudt.edu.cn,}} \\
  \small{\texttt{liu.hd@sz.tsinghua.edu.cn}} \\
  \textsuperscript{*}Equal contribution, \textsuperscript{$\dagger$}Corresponding authors
}

\begin{document}

\maketitle

\begin{abstract}
Temporal Knowledge Graph Question Answering (TKGQA) is challenging because it requires multi-hop reasoning under complex temporal constraints. Recent LLM-based approaches have improved semantic modeling for this task, but many still rely on fixed reasoning workflows or costly post-training, which can limit adaptability and make error recovery difficult. We show that \emph{enabling an off-the-shelf Large Language Model (LLM) to determine its next action} is already effective in a zero-shot setting. Based on this insight, we propose \textbf{AT2QA}, an \textbf{A}utonomous and \textbf{T}raining-free \textbf{A}gent for \textbf{T}KG \textbf{Q}uestion \textbf{A}nswering. AT2QA empowers the LLM to iteratively interact with the TKG via a generic search tool, inherently enabling autonomous exploration and dynamic self-correction during reasoning. To further elicit the LLM's potential for complex temporal reasoning, we introduce a training-free experience mining mechanism that distills a compact few-shot demonstration library from successful self-generated trajectories. AT2QA also yields a transparent audit trail for every prediction. Experiments on three challenging benchmarks—MultiTQ, Timeline-CronQuestion, and Timeline-ICEWS-Actor—show that AT2QA achieves new state-of-the-art performance, surpassing the strongest baselines by 10.7, 4.9, and 11.2 absolute points, respectively.  Our code is available at \href{https://anonymous.4open.science/r/AT2QA-Official-Code-7DE8/}{[Anonymous GitHub]}.

\end{abstract}


\section{Introduction}
While traditional Knowledge Graphs (KGs) have long served as a fundamental infrastructure for question answering, real-world facts are inherently dynamic and time-dependent. To prevent models from relying on outdated knowledge, Temporal Knowledge Graphs (TKGs) capture this evolution by extending static facts into quadruples denoted as $<subject, relation, object, timestamp>$ \cite{saxena2021question}. Consequently, Temporal Knowledge Graph Question Answering (TKGQA) is sub
\begin{center}
  \includegraphics[
    width=\linewidth,
    trim=0.5 31.5 0 32.5,
    clip
  ]{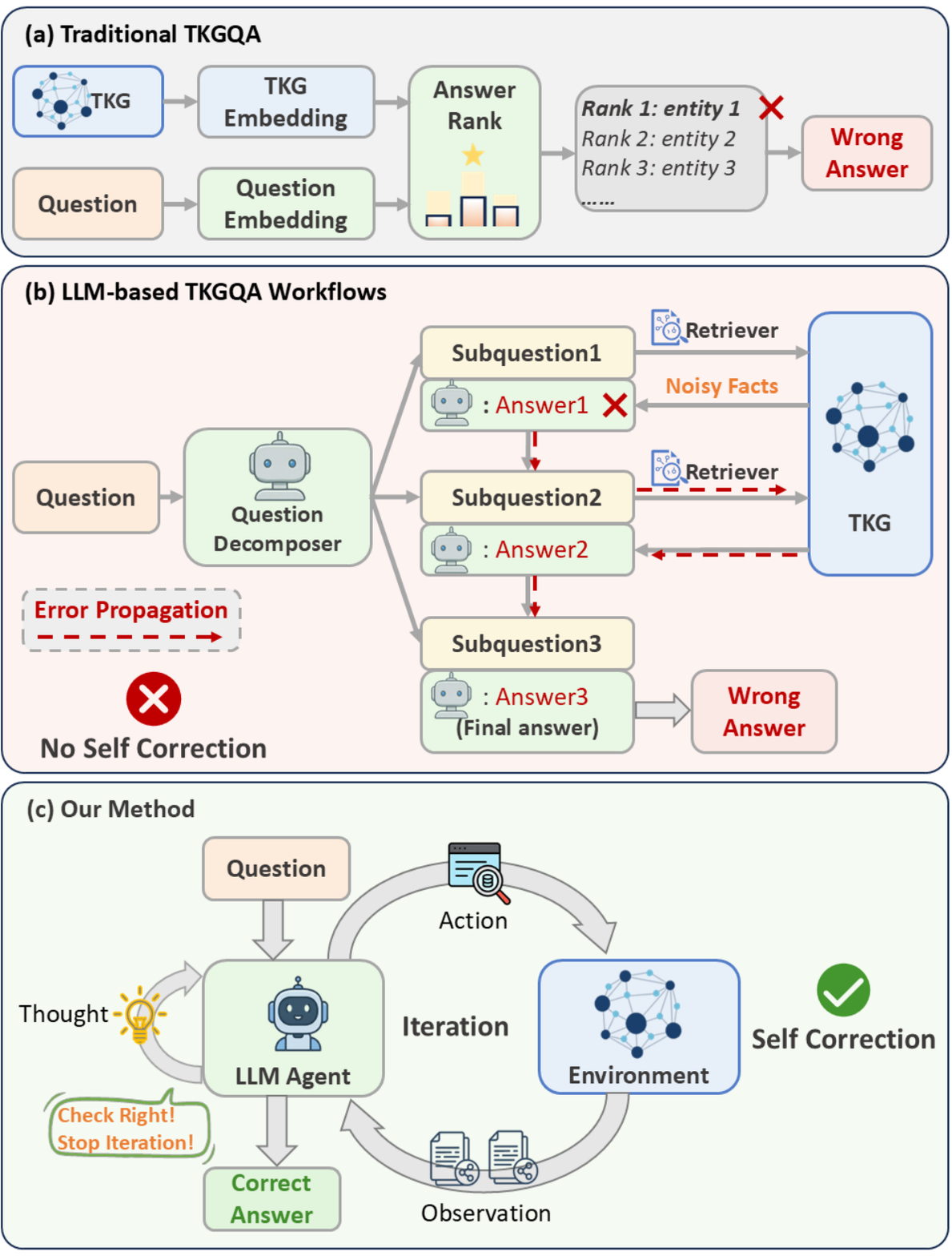}
  {\captionsetup{hypcap=false}
  \captionof{figure}{Comparison between \textbf{AT2QA} and existing methods. \textbf{(a) Embedding-based Methods} rely on vector representations, lacking semantic understanding. \textbf{(b) LLM-based Workflows} execute fixed pipelines, making them vulnerable to cascading errors. \textbf{(c)} \textbf{Our AT2QA} operates as an autonomous agent. It autonomously explores and self-corrects its reasoning trajectory by iteratively interacting with the TKG.}}
  \label{fig:comparison}
\end{center}
\medskip
stantially more challenging than conventional static Knowledge Graph Question Answering (KGQA). This is because answering complex temporal questions often requires multi-hop reasoning over structural entities while simultaneously satisfying combined or multi-granular temporal constraints.

In recent years, TKGQA has shifted from traditional task-specific architectures to Large Language Model (LLM)-based frameworks. Traditional embedding-based methods \cite{mavromatis2022tempoqr, chen2023multi} map questions and TKG facts into low-dimensional vector spaces and rank candidate answers with scoring functions (as illustrated in Figure~\ref{fig:comparison}(a)), but they often struggle to capture the complex semantics of temporal constraints in natural language questions \cite{chen2023multi}. In contrast, LLMs have shown strong performance on complex natural language tasks \cite{yang2025, peng2025}. Consequently, recent studies have increasingly explored the potential of LLMs for TKGQA \cite{chen2024ari, qian2024timer4, gong2025rtqa, qian2025pok}. These methods typically incorporate dynamic graph knowledge into LLMs through retrieval-augmented generation or fine-tuning under predefined workflows. Nevertheless, existing approaches still exhibit three key limitations:

\textbf{(1) Rigid Workflows and Cascading Errors.} Existing methods typically constrain LLMs within fixed, human-designed reasoning pipelines rather than allowing them to interact with the TKG environment to determine their next actions. This rigidity limits the model’s ability to autonomously explore its reasoning trajectory and dynamically self-correct during execution. As illustrated in Figure~\ref{fig:comparison}(b), when strictly following a predefined decomposition workflow, an initial retrieval failure inevitably cascades through subsequent steps, amplifying the error and ultimately yielding an incorrect final answer.


\textbf{(2) Prohibitive Training Cost and Limited Generalization.} 
To bridge the mismatch between LLMs and TKGs, prior work largely depends on costly post-training, ranging from SFT-based methods \cite{gao2024two,qian2024timer4,qian2025pok} to recent RL-based agents such as Temp-R1 and TKG-Thinker, the latter even requiring a two-stage SFT+RL pipeline \cite{gong2026tempr1,jiang2026tkg}. However, RL training is widely recognized as computationally costly and complex \cite{sidahmed2024parameterefficientreinforcementlearning}. Beyond the hardware burden, graph-specific post-training can also reduce plug-and-play transfer to unseen or dynamically evolving TKGs.

\textbf{(3) Limited Interpretability.} While recent LLM-based methods offer intermediate reasoning chains as explanations \cite{gong2025rtqa, qian2025pok}, their rigid pipelines separate reasoning from dynamic graph exploration. As a result, when early retrieval steps fail, these explanations can become ungrounded and susceptible to hallucination. Without a verifiable audit trail, it is difficult to determine whether failures arise from flawed reasoning or incorrect retrieved evidence.

To overcome these limitations, we propose \textbf{AT2QA} (\textbf{A}utonomous and \textbf{T}raining-free \textbf{A}gent for \textbf{T}KG \textbf{Q}uestion \textbf{A}nswering), a framework that enables autonomous temporal reasoning without parameter updates. As shown in Figure~\ref{fig:comparison}(c), AT2QA equips an LLM with a generic search tool and allows it to interact with the TKG environment in an iterative manner. Instead of following a fixed workflow, the model can iteratively verify retrieved evidence and reformulate queries when current evidence is insufficient or contradictory, thereby improving recovery from intermediate errors in multi-step reasoning.

To make such autonomous reasoning more reliable on complex queries, AT2QA further introduces a training-free experience mining mechanism. Using rule-based rewards, it extracts high-quality demonstration trajectories from the model's own successful explorations and uses them as few-shot guidance, without any parameter updates. In addition, because each internal \texttt{<think>} process, autonomous \texttt{<search>} action, and environmental \texttt{<observation>} is logged, AT2QA provides an explicit and auditable evidence chain for each prediction. Our main contributions are summarized as follows:

\textbf{(1)An Autonomous Agent Framework for TKGQA:} We introduce AT2QA, a novel autonomy-first agent framework. By enabling the LLM to iteratively interact with the TKG environment, our method inherently enables dynamic self-correction and mitigates cascading errors in complex temporal reasoning.

\textbf{(2)Training-Free Experience Mining:} We propose a highly efficient experience mining strategy. By distilling a compact few-shot library from the model's self-generated successful trajectories, this mechanism further enhances the model's potential for complex temporal reasoning without any parameter updates and facilitates broad plug-and-play generalization.

\textbf{(3)State-of-the-Art Performance and Traceability:} AT2QA achieves state-of-the-art performance on three challenging TKGQA benchmarks (MultiTQ, Timeline-CronQuestion, and Timeline-ICEWS-Actor) with Hits@1 scores of 88.7\%, 75.4\%, and 75.4\%, outperforming the best baselines by 10.7, 4.9, and 11.2 absolute points, respectively. AT2QA also yields a transparent and verifiable audit trail for every prediction.

\section{Related Work}

\subsection{Traditional TKGQA}
Traditional Temporal knowledge graph question answering (TKGQA) primarily relied on representation learning and logical parsing. Embedding-based methods \cite{saxena2021question, mavromatis2022tempoqr, chen2023multi} and approaches like TSQA \cite{shang2022tsqa}, which formulate the task as temporal knowledge graph completion, encode entities and temporal relations into low-dimensional spaces and rely on scoring functions to evaluate the plausibility of candidate facts. Although these methods laid a crucial foundation for the field, traditional embedding representations often act as opaque "black boxes" with weak system interpretability \cite{cai2023temporal}. In contrast, semantic parsing-based methods \cite{jia2018tequila, neelam2021semantic, Ding2022, chen2024progtqa} attempt to translate natural language questions into logical query expressions, while Graph Neural Networks (GNNs) \cite{jia2021complex, mavromatis-etal-2023-twirgcn, liu2023} have been introduced to capture the complex structural dependencies within the graphs. These traditional paradigms suffer from a common bottleneck: they typically demand prohibitive resources for specialized training, making it exceedingly difficult to generalize to unseen TKGs.

\subsection{LLM-based TKGQA Workflows}
In recent years, the introduction of Large Language Models has profoundly driven a paradigm shift in TKGQA, rapidly steering the research focus toward leveraging the powerful in-context learning and semantic parsing capabilities of LLMs for RAG or interactive querying over TKGs. Existing LLM integration methods primarily focus on RAG and prompt engineering: ARI \cite{chen2024ari} enhances the temporal adaptability of models via time-aware training signals; $\text{TimeR}^{4}$ \cite{qian2024timer4} and PoK \cite{qian2025pok} generate more comprehensive reasoning plans by improving retrieval components; and TempAgent \cite{qianyihu-etal-2025-time} adapts the ReAct paradigm to the temporal domain by designing a toolkit with 10 specific temporal tools. RTQA \cite{gong2025rtqa} adopts a bottom-up decomposition strategy to solve sub-questions recursively, while MemoTime \cite{tan2025memotime} utilizes closed-source APIs for reasoning and stores solution paths as memories. Nevertheless, these approaches either rely on rigid, human-designed hard-coded paths, thereby constraining LLMs of their intrinsic global planning and autonomous self-correction capabilities, or necessitate prohibitively expensive and time-consuming SFT training. Recently, Temp-R1 \cite{gong2026tempr1} and TKG-Thinker \cite{jiang2026tkg} have explored the use of reinforcement learning to equip agents with temporal reasoning capabilities. Despite their contributions to optimization efficiency, these methods still fundamentally depend on updating model parameters.

\section{Intuition}
\label{sec:intuition}

Before presenting our methodology AT2QA, we detail two pivotal empirical observations that establish the foundation of our framework. These findings challenge the prevailing assumption in TKGQA that high performance necessitates either complex supervised fine-tuning or rigid, human-crafted reasoning workflows.

\subsection{The Language Model is Smart Enough to Decide What to Do Next}
\label{subsec:autonomy}

Recent research in TKGQA typically constrains LLMs within static decomposition frameworks or predefined reasoning paths\cite{gong2026tempr1,jiang2026tkg}. Such approaches implicitly assume that LLMs lack the capability to independently navigate the intricate temporal constraints and structural dependencies of knowledge graphs\cite{qian2025pok}. However, our preliminary experiments suggest a contrary conclusion.

\begin{figure}[t]
    \centering
    \includegraphics[width=1.0\linewidth]{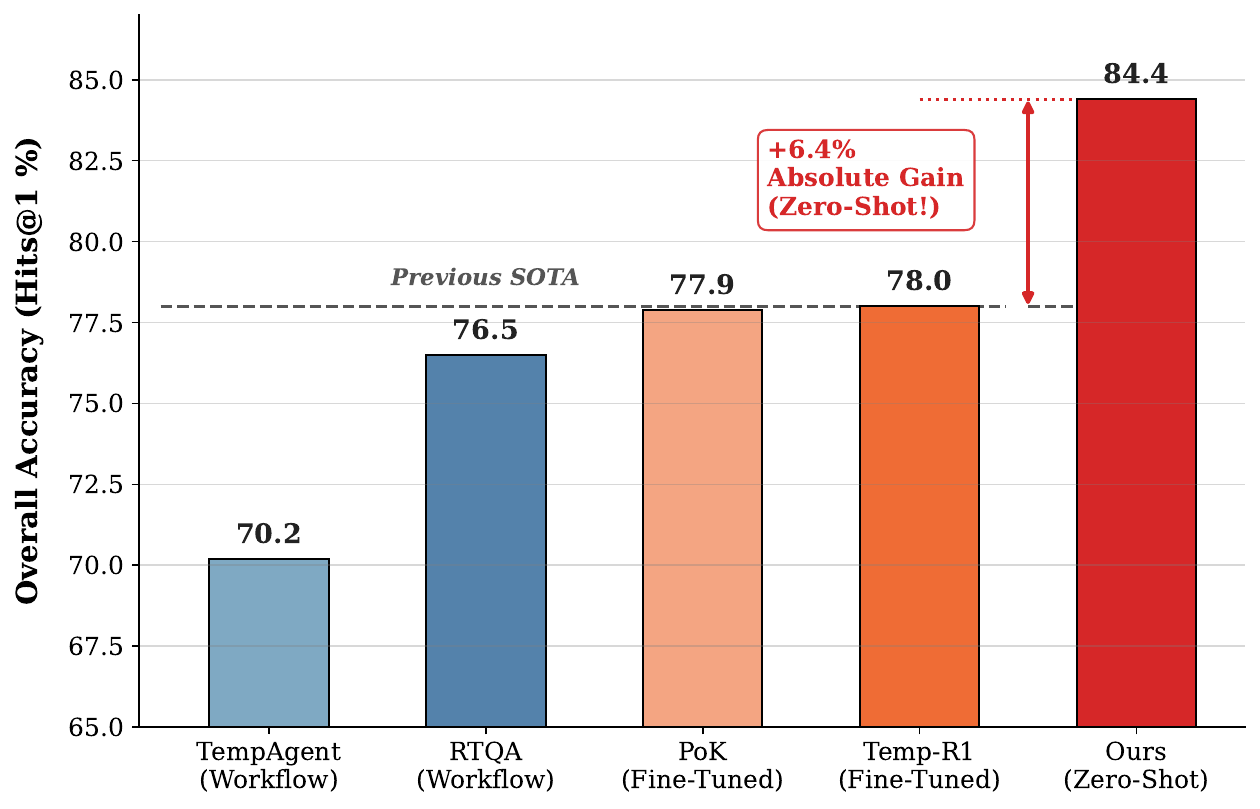}
    \caption{Performance comparison on the MultiTQ benchmark. In a zero-shot setting, the autonomous LLM agent surpasses existing baselines, highlighting the efficacy of unlocking the model's inherent decision-making capabilities.}
    \label{fig:comparison_sota}
\end{figure}

We observe that when an off-the-shelf LLM is equipped with a generic \texttt{search} tool and granted the \textit{autonomy} to determine when and what to retrieve, it exhibits strong planning proficiency. As illustrated in Figure~\ref{fig:comparison_sota}, even in a \textbf{zero-shot setting without any parameter updates}, AT2QA significantly outperforms current state-of-the-art (SOTA) methods that rely on extensive supervised training or rigid, human-crafted reasoning workflows. This phenomenon indicates that modern LLMs already possess the intrinsic intelligence required to solve complex temporal queries. \textbf{The bottleneck lies not in the model's reasoning capacity, but in the lack of an appropriate interface that allows the model to exercise its autonomy for information retrieval}.

\subsection{Eliciting Capabilities Instead of Fine-Tuning}
\label{subsec:no_finetune}

The second observation addresses the necessity of computational heavy-lifting. While Supervised Fine-Tuning is a standard paradigm to align models with TKG tasks, we hypothesize that the requisite reasoning patterns are already latent within the pre-trained weights of strong LLMs.

\begin{figure}[t]
    \centering
    \includegraphics[width=1.0\linewidth]{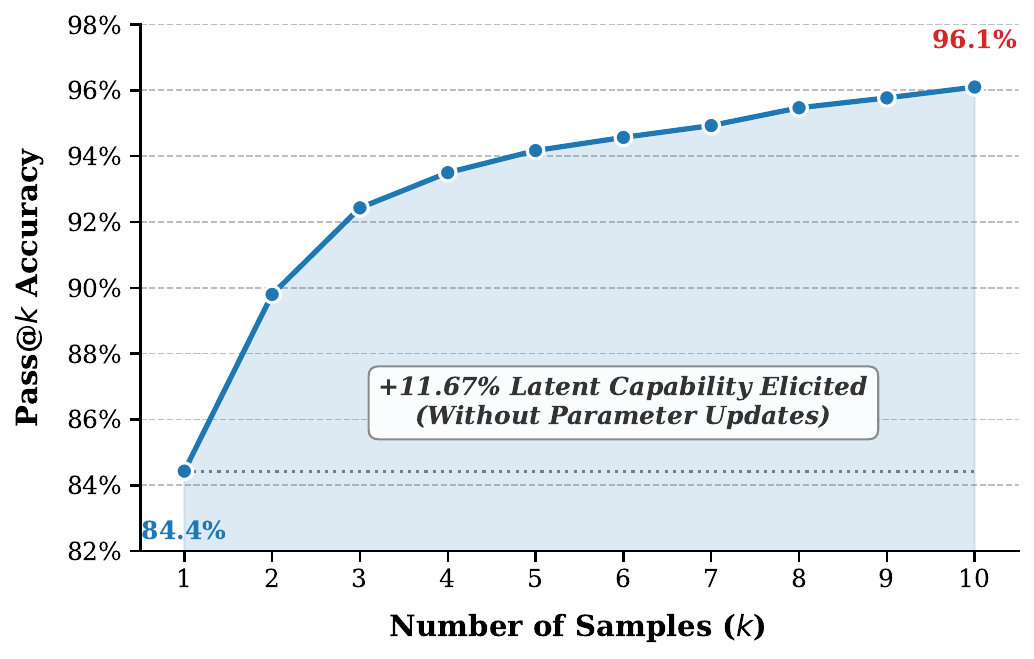}
    \caption{Pass@k analysis of our method. In a zero-shot setting, our method achieves >84\% Pass@1 accuracy. For difficult queries that initially fail, repeated sampling ($k=10$) successfully retrieves the correct answer, suggesting the reasoning capability is present but dormant.}
    \label{fig:pass_at_k}
\end{figure}

To validate this, we conducted a \textit{Pass@k} analysis on a randomly sampled subset of 3,000 questions from the MultiTQ benchmark. Specifically, the \textit{Pass@k} metric evaluates whether at least one out of $k$ independent generation attempts yields the correct answer\cite{yue2025doesreinforcementlearningreally,lv2025hiddenlinkrlhfcontrastive}. As depicted in Figure~\ref{fig:pass_at_k}, AT2QA achieves an impressive accuracy of over \textbf{84\%} at \textit{Pass@1} in a zero-shot setting. Crucially, for the subset of "hard" queries where the model initially failed, we observed that repeated sampling (increasing $k$) rapidly closes the performance gap. At $k=10$, the model is able to generate at least one correct reasoning path for nearly all queries.

This finding indicates that \textbf{the model inherently possesses the capability to solve even the most challenging temporal reasoning problems.} Empirical evidence proves that the correct reasoning trajectory \textbf{already exists} within the model's latent space; the model effectively ``knows'' the solution but failed to assign the highest probability to the optimal path in a single inference step. Crucially, the observation that the model converges to the correct answer with a minimal number of trials ($k<10$) strongly suggests that computationally expensive fine-tuning may be avoidable in TKGQA. Such a low $k$ threshold implies that the reasoning capability is easily accessible and can likely be deterministically activated solely through appropriate prompting strategies\cite{snell2024scalingllmtesttimecompute,wang2023selfconsistencyimproveschainthought}. Consequently, this perspective suggests shifts from \textit{injecting} new capabilities via parameter updates to \textit{eliciting} dormant capabilities via optimal prompting. \textbf{Our problem thus evolves into identifying the most effective prompts}—synthesizing experience from successful samples—to stably trigger the LLM's latent potential without the need for gradient updates.


\section{Methodology}
\label{sec:method}

\begin{figure*}[t]
  \centering
  \includegraphics[
    width=\textwidth,
    trim=53 81 60 83,
    clip
  ]{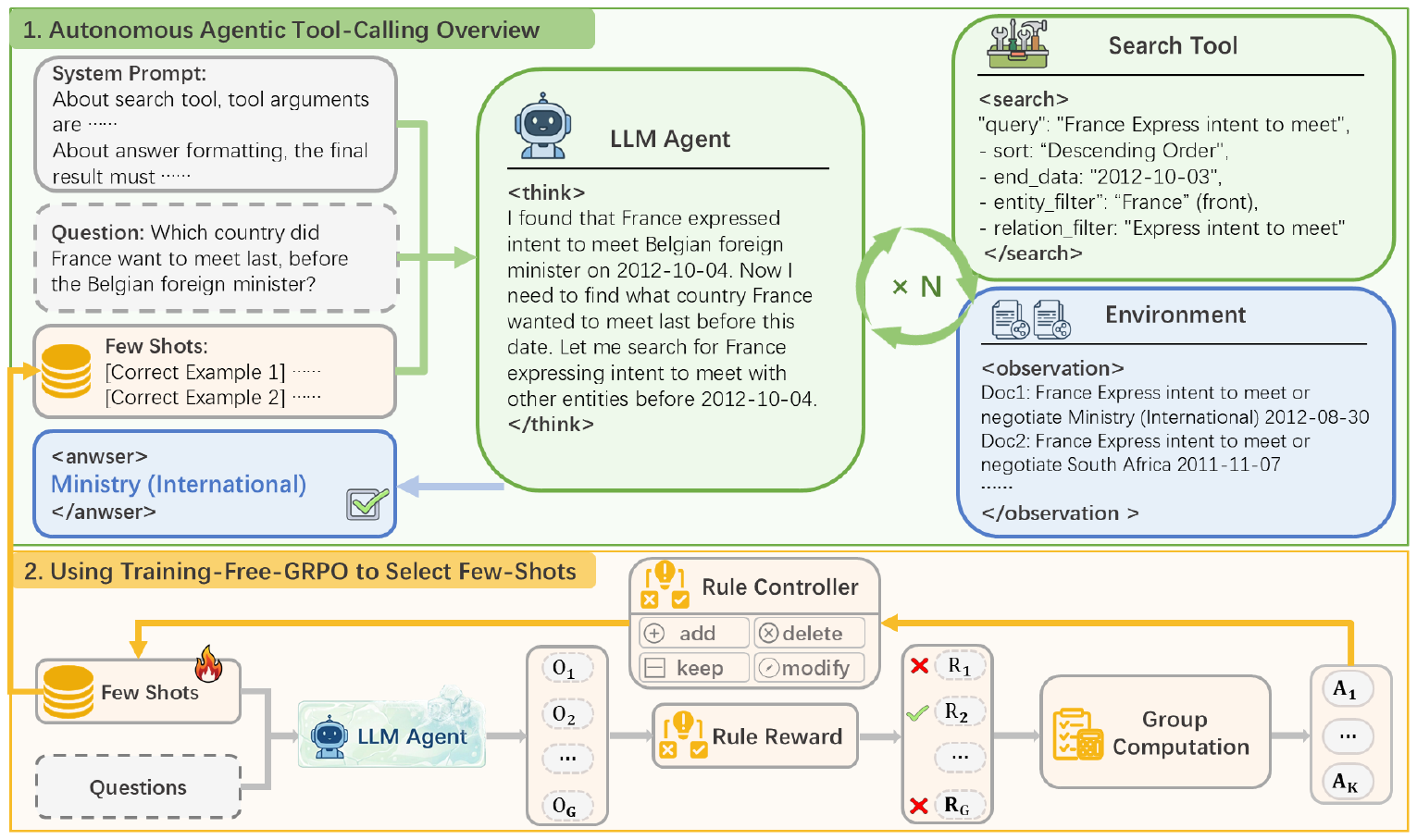}
  \caption{The overview of our proposed framework AT2QA. Top: At inference, an LLM agent repeatedly queries a \texttt{Search} tool to interact with the TKG environment until sufficient evidence is collected. Inputs include a system prompt, the question, and few-shot demonstrations. Bottom: The few-shot library is selected from candidate trajectories via training-free GRPO-style rule editing with rule-based rewards.}
  \label{fig:method}
  \vspace{-0.2cm}
\end{figure*}

In this section, we introduce AT2QA, a fully autonomous and training-free LLM agent framework capable of dynamic self-correction for TKGQA Question Answering. AT2QA consists of two core components, as shown in Figure \ref{fig:method}: a \textit{retrieval-augmented reasoning agent} equipped with a structured temporal search tool, and a \textit{trajectory optimization strategy} that mines effective few-shot demonstrations from the model's self-generated experiences.

\subsection{Preliminaries and Problem Formulation}
A Temporal Knowledge Graph (TKG) is defined as a collection of quadruples $\mathcal{G} = \{(e_h, r, e_l, \tau)\} \subseteq \mathcal{E} \times \mathcal{R} \times \mathcal{E} \times \mathcal{T}$, where $\mathcal{E}$, $\mathcal{R}$, and $\mathcal{T}$ denote the sets of entities, relations, and timestamps, respectively.
Given a natural language question $Q$, the goal is to derive the correct answer $y$ by reasoning over the facts in $\mathcal{G}$. Unlike static QA, $Q$ typically contains implicit or explicit temporal constraints that require filtering facts based on $\tau$.

\subsection{Tool-Augmented Reasoning Framework}
To bridge the gap between the LLM's parametric knowledge and the structured data in $\mathcal{G}$, we develop a \texttt{Search} tool that supports both semantic retrieval and symbolic filtering.

\paragraph{Structured Temporal Retrieval.}
The \texttt{Search} tool accepts a query string $q$ along with a set of structured constraints $C = \{c_{time}, c_{entity}, c_{rel}\}$.
\begin{itemize}
    \item \textbf{Filtering:} Before semantic matching, the tool filters $\mathcal{G}$ to obtain a candidate set $\mathcal{G}_{sub} \subset \mathcal{G}$. Constraints act as a conjunctive filter: $c_{time}$ specifies an inclusive time window $[\tau_{start}, \tau_{end}]$; $c_{entity}$ restricts involved entities (e.g., head/tail candidates); and $c_{rel}$ filters for exact relation matches. This ensures that retrieval is grounded in valid temporal and structural contexts.
    \item \textbf{Semantic Ranking:} We employ a dense retrieval approach. Each fact $f \in \mathcal{G}_{sub}$ and the query $q$ are encoded into a shared embedding space. Facts are ranked based on the cosine similarity score $s(f, q) = \cos(enc(f), enc(q))$.
    \item \textbf{Hybrid Sorting:} To handle temporal nuances, the tool supports two sorting modes: (1) \textit{Relevance-based}, sorting solely by $s(f, q)$, and (2) \textit{Time-based}, where the top-$m$ relevant facts are re-ranked chronologically. This exposes the temporal evolution of facts to the agent.
\end{itemize}

\paragraph{Reasoning Process.}
The LLM operates as an autonomous agent. At each step $t$, it generates a thought and a tool call action $a_t = \texttt{Search}(q_t, C_t)$ based on the history of previous actions and observations. The iterative loop continues until the agent generates a special termination token or reaches the maximum step limit $T_{max}$. This multi-turn interaction allows the agent to perform \textit{Self-Correction}: if retrieved evidence conflicts with the hypothesis, the agent can refine its constraints $C_{t+1}$ (e.g., expanding the time window) in subsequent turns.

\subsection{Training-Free Experience Mining}
\label{subsec:mining}
To enhance the agent without any parameter updates, we adopt a \textbf{Training-Free GRPO-style} experience selection scheme~\cite{cai2025trainingfreegrouprelativepolicy}. 
Instead of optimizing model weights, we generate a pool of candidate reasoning traces and distill a small set of \emph{high-value} ``advantage experiences'' that most effectively improve the LLM when used as few-shot demonstrations.

\paragraph{Trajectory Sampling.}
Given a mini-batch of $N$ training questions and a group size $g$, we sample $G$ interaction traces per question using stochastic decoding, resulting in $N\times G$ traces in total.
For a question $Q_i$, the sampled group is denoted as
$\mathcal{O}_i=\{O_{i,1},\dots,O_{i,G}\}$, where each $O_{i,j}$ contains the full tool-interaction transcript and a final predicted answer.

\paragraph{Rule-Based Reward.}
We assign each trace a binary rule reward $R_{i,j}\in\{0,1\}$ by exact match:
\[
R_{i,j}=\mathbf{1}\!\left[\hat{y}_{i,j}=y_i^*\right],
\]
where $\hat{y}_{i,j}$ is the answer extracted from $O_{i,j}$ and $y_i^*$ is the gold answer for $Q_i$.
We retain only successful traces
$\mathcal{O}^{+}_i=\{O_{i,j}\in\mathcal{O}_i \mid R_{i,j}=1\}$.

\paragraph{LLM-Guided Group Computation.}
Among correct traces $\mathcal{O}^{+}_i$, demonstrations vary in usefulness. We thus let the LLM rank them by \emph{marginal instructional value} (i.e., how much they provide new, ``aha''-style guidance beyond what the model already knows). The top-ranked traces are distilled into \emph{advantage experience texts} $\{A_{i,1},\dots,A_{i,K}\}$.

\paragraph{Experience Library Rule-Controller.}
We then validate each advantage candidate by measuring the validation-set gain after adding it to the current library. With a fixed library budget of $K$ shots, we keep the $K$ experiences that yield the largest validation improvements (from both existing and newly mined candidates), forming the final library $\mathcal{D}_{demo}$ for test-time inference.

\section{Experiments}
\label{sec:experiments}

\subsection{Experimental Setup}
\label{subsec:setup}

\paragraph{Datasets.}
We evaluate AT2QA on two challenging TKGQA benchmarks: MultiTQ \cite{chen2023multi} and TimelineKGQA \cite{timelinekgqa}, which comprises Timeline-CronQuestion and Timeline-ICEWS-Actor. These datasets collectively assess diverse reasoning capabilities, spanning multi-granular timestamp constraints, chronological event tracking, and actor-centric temporal dynamics. Detailed statistics and metrics are deferred to Appendix ~\ref{sec:appendix-Dataset}.

\paragraph{Baselines.}
We compare AT2QA against strong baselines from two paradigms of TKGQA: \textbf{(1) TKG embedding-based methods}, including EmbedKGQA \cite{Jin_2022}, CronKGQA \cite{saxena2021question}, and MultiQA \cite{chen2023multi}; and \textbf{(2) LLM-based methods}, including prompt-based workflows such as ARI \cite{chen2024ari}, TempAgent \cite{qianyihu-etal-2025-time}, MemoTime \cite{tan2025memotime}, and RTQA \cite{gong2025rtqa}, as well as training-based approaches such as Search-R1 \cite{jin2025searchr1}, TimeR$^4$ \cite{qian2024timer4}, PoK \cite{qian2025pok}, and Temp-R1 \cite{gong2026tempr1}. These baselines cover both traditional TKGQA systems and recent LLM-centered paradigms, providing a comprehensive comparison against prior state-of-the-art methods. Following standard practice in TKGQA, we report the \textbf{Hits@1} (Exact Match) metric for all methods.



\subsection{Implementation Details}
\label{subsec:implementation}

AT2QA uses DeepSeek-V3.2 with the server default decoding configuration (temperature $=1.0$). All facts are embedded offline by GLM-Embedding-3 (256-d), and the search tool performs brute-force cosine-similarity retrieval over structurally filtered candidates, returning at most $10$ facts per call. We cap the maximum interaction rounds at $T_{\max}=20$ and use a fixed library of $K=3$ demonstrations for training-free optimization. For ablation, we reimplemented RTQA in an OpenAI-compatible tool-calling form while preserving its rigid workflow, and replaced its original retriever with our advanced search tool to obtain the ``RTQA + advanced tool'' variant, enabling controlled comparison under the same tool interface. At official API rates, AT2QA costs under \$150 on MultiTQ and under \$50 on TimelineKGQA, while training-free GRPO costs under \$7 per dataset..

\subsection{Main Results}
\label{subsec:main_results}

\begin{table*}[htbp]
\centering
\small 
\setlength{\tabcolsep}{8.5pt} 
\renewcommand{\arraystretch}{1.1} 

\centering
\begin{tabular}{l cccc c cccc}
\toprule
\multirow{2}{*}{\textbf{Method}} & \multicolumn{4}{c}{\textbf{TimelineKGQA-CronQuestion}} & & \multicolumn{4}{c}{\textbf{TimelineKGQA-ICEWS-Actor}} \\
\cmidrule{2-5} \cmidrule{7-10}
& Overall & Simple & Medium & Complex & & Overall & Simple & Medium & Complex \\
\midrule
RAG Baseline & 0.235 & 0.704 & 0.092 & 0.009 & & 0.265 & 0.660 & 0.128 & 0.011 \\
LLaMA2-7B    & 0.169 & 0.049 & 0.143 & 0.282 & & 0.111 & 0.035 & 0.066 & 0.322 \\
GPT-4o       & 0.206 & 0.069 & 0.130 & 0.376 & & 0.113 & 0.051 & 0.035 & 0.353 \\
RTQA         & 0.298 & 0.608 & 0.218 & 0.135 & & --    & --    & --    & --    \\
PoK          & 0.651 & 0.737 & \underline{0.539} & \underline{0.683} & & 0.602 & 0.744 & \underline{0.456} & 0.578 \\
Temp-R1 & \underline{0.705} & \textbf{0.960} & 0.486 & 0.672 & & \underline{0.642} & \textbf{0.866} & 0.388 & \underline{0.595} \\
\midrule
\rowcolor{highlightrow}
\textbf{AT2QA} & \textbf{0.754} & \underline{0.831} & \textbf{0.631} & \textbf{0.803} & & \textbf{0.754} & \underline{0.859} & \textbf{0.627} & \textbf{0.768} \\
\bottomrule
\end{tabular}
\caption{Performance comparison on TimelineKGQA-CronQuestion and TimelineKGQA-ICEWS-Actor datasets. \textbf{Bold} indicates the best performance, and \underline{underline} indicates the second best.}
\label{tab:TimeLineKGQA_performance}
\end{table*}

\begin{table}[t]
\centering
\small 
\setlength{\tabcolsep}{3.5pt} 
\renewcommand{\arraystretch}{1.1} 

\begin{tabular}{lcccccc}
\toprule
\multirow{2}{*}{\textbf{Method}} & \multirow{2}{*}{\textbf{Overall}} & \multicolumn{2}{c}{\textbf{Question Type}} & & \multicolumn{2}{c}{\textbf{Answer Type}} \\
\cline{3-4} \cline{6-7}
& & \textbf{multiple} & \textbf{single} & & \textbf{entity} & \textbf{time} \\
\midrule
\multicolumn{7}{c}{\textbf{TKG Embedding-based Methods}} \\
\midrule
EmbedKGQA & 0.206 & 0.134 & 0.235 & & 0.290 & 0.001 \\
CronKGQA  & 0.279 & 0.134 & 0.337 & & 0.328 & 0.156 \\
MultiQA   & 0.293 & 0.159 & 0.347 & & 0.349 & 0.157 \\
\midrule
\multicolumn{7}{c}{\textbf{LLM-based Static Workflows}} \\
\midrule
Search-R1 & 0.352 & 0.094 & 0.474 & & 0.230 & 0.705 \\
ARI       & 0.380 & 0.210 & 0.680 & & 0.394 & 0.344 \\
TempAgent & 0.702 & 0.316 & 0.857 & & 0.624 & 0.870 \\
TimeR$^4$ & 0.728 & 0.335 & 0.887 & & 0.639 & 0.945 \\
MemoTime  & 0.730 & 0.459 & 0.829 & & 0.677 & 0.846 \\
RTQA      & 0.765 & 0.424 & 0.902 & & 0.692 & 0.942 \\
PoK       & 0.779 & 0.409 & \underline{0.929} & & 0.696 & \underline{0.962} \\
Temp-R1   & \underline{0.780} & \underline{0.550} & 0.888 & & \underline{0.714} & \textbf{0.969} \\
\midrule
\multicolumn{7}{c}{\textbf{Ours (Autonomous Training-free Agent)}} \\
\midrule
\rowcolor{highlightrow}
\textbf{AT2QA} & \textbf{0.887} & \textbf{0.751} & \textbf{0.942} & & \textbf{0.864} & 0.945 \\
\bottomrule
\end{tabular}
\caption{Performance comparison on the MultiTQ test set.\textbf{Bold} indicates the best performance, and \underline{underline} indicates the second best.}
\label{tab:main_results}
\vspace{-0.4cm}
\end{table}
The main results on MultiTQ are shown in Table~\ref{tab:main_results}. 
AT2QA achieves a new state-of-the-art with an overall accuracy of \textbf{88.7\%}, outperforming the previous best model, Temp-R1 (78.0\%), by \textbf{10.7} points. 
The advantage is most pronounced on \textbf{multiple}-answer questions, where AT2QA reaches \textbf{75.1\%}, exceeding the previous best result (55.0\%) by \textbf{20.1} points, which highlights its strength in exhaustive temporal multi-hop reasoning. 
AT2QA also achieves the best performance on \textbf{entity} answers (86.4\%) and remains competitive on \textbf{time} answers (94.5\%), falling only slightly below Temp-R1. 
Overall, these results show that AT2QA is highly effective for complex TKGQA.

To further evaluate generalization, we test AT2QA on TimelineKGQA-CronQuestion and TimelineKGQA-ICEWS-Actor (Table~\ref{tab:TimeLineKGQA_performance}). AT2QA achieves the best overall accuracy on both datasets, reaching \textbf{75.4\%} on each and surpassing Temp-R1 by \textbf{4.9} and \textbf{11.2} points, respectively. Its gains are concentrated on the \textbf{medium} and \textbf{complex} subsets, where it consistently outperforms prior methods, while Temp-R1 remains slightly stronger on \textbf{simple} questions; part of this apparent gap is due to benchmark undercounting issues (Appendix ~\ref{sec:appendix-TimelineCron}). This suggests that AT2QA is particularly effective on harder TKGQA cases requiring adaptive search, multi-step evidence accumulation, and dynamic reasoning.

\subsection{Ablation Study}
\label{subsec:ablation}

We conduct ablations to answer four questions: 
(1) whether AT2QA improves simply by allowing more interaction rounds, 
(2) whether the main gains come from better tools or from autonomy, 
(3) how much training-free GRPO-selected few-shot demonstrations contribute, and 
(4) how sensitive the framework is to retrieval depth and backbone choice.

\subsubsection{Effect of the Interaction Budget}

\begin{figure}[t]
    \centering
    \includegraphics[width=\linewidth]{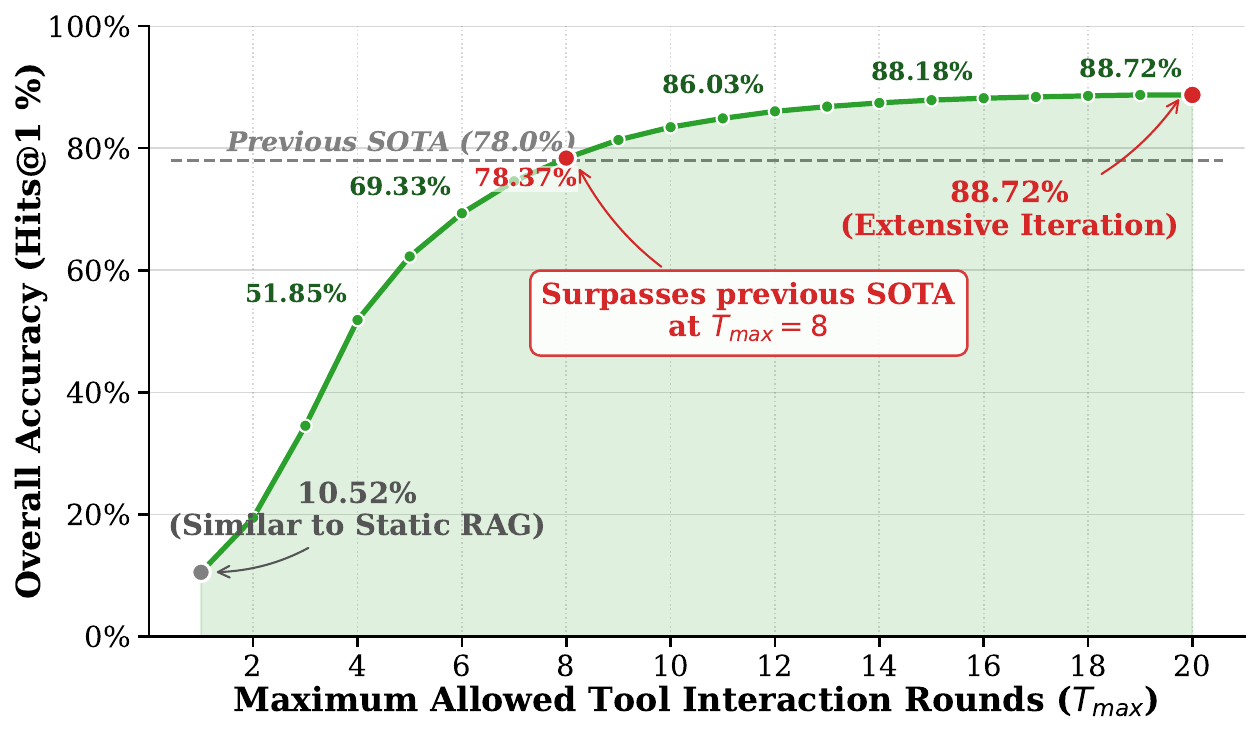}
    \caption{Effect of the maximum interaction rounds $T_{\max}$ on MultiTQ. 
    AT2QA reaches most of its gain with a moderate interaction budget, rather than relying on very deep search.}
    \label{fig:interaction_rounds}
    \vspace{-0.4cm}
\end{figure}

A natural concern is that the improvement comes from allowing many search attempts. 
Figure~\ref{fig:interaction_rounds} shows that this is not the case. 
Performance increases rapidly as the maximum interaction budget grows, indicating that multi-turn interaction is important; however, the gain does not rely on very deep search. 
AT2QA already surpasses previous SOTA at $T_{\max}=8$, after which the curve shows clear diminishing returns, and the best result is reached at around 19 rounds. 
We therefore use $T_{\max}=20$ in the final system.

\subsubsection{Main Sources of Improvement}

Table~\ref{tab:main_ablation} isolates the contributions of the tool, the reasoning paradigm, and the training-free GRPO-selected few-shot demonstrations.
Under the rigid RTQA workflow, replacing the base tool with our advanced tool brings only a marginal gain from 76.2\% to 76.7\%, showing that the tool itself is not the main source of improvement. 
With the same advanced tool, our autonomous agent reaches 84.4\% in the zero-shot setting, a \textbf{+7.7 point gain} over rigid RTQA, indicating \textbf{that the major gain comes from autonomy}. 
On top of that, adding the GRPO-selected few-shot demonstrations further improves performance from \textbf{84.4\% to 88.7\% (+4.3)}, showing that \textbf{training-free GRPO provides another substantial boost}.

We also ablate the components of the advanced tool in the zero-shot autonomous setting. 
Temporal modules contribute the most: time limit and time-aware sorting improve accuracy to 81.7\% and 81.4\%, while entity and relation filtering yield smaller gains (79.5\% and 79.9\%). 
Combining all components reaches 84.4\%, suggesting that \textbf{the advanced tool offers auxiliary gains}.

\begin{table}[t]
\centering
\footnotesize
\setlength{\tabcolsep}{2.5pt}
\renewcommand{\arraystretch}{1.05}

\begin{tabularx}{\columnwidth}{@{}p{0.25\columnwidth} >{\raggedright\arraybackslash}X >{\centering\arraybackslash}p{0.12\columnwidth}@{}}
\toprule
\textbf{Paradigm} & \textbf{Configuration} & \textbf{Hits@1}\\
\midrule
LLM only & DeepSeek V3.2 without search tool & 0.100 \\
\midrule
\multirow[t]{2}{*}{\makecell[tl]{Rigid workflow \\ (RTQA)}}
& + Base tool (semantic relevance) & 0.762 \\
& + Advanced tool (all components) & 0.767 \\
\midrule
\multirow[t]{6}{*}{\makecell[tl]{Autonomous \\ agent \\ (zero-shot)}}
& + Base tool (semantic relevance) & 0.791 \\
& + Base tool + time limit & 0.817 \\
& + Base tool + time-aware sorting & 0.814 \\
& + Base tool + entity filter & 0.795 \\
& + Base tool + relation filter & 0.799 \\
& + Advanced tool (all components) & 0.844 \\
\midrule
\makecell[l]{Autonomous \\ agent+ training\\free GRPO}
& \makecell[l]{Advanced tool + \\ selected few-shot\\ demonstrations}
& \multicolumn{1}{c}{\textbf{0.887}} \\
\bottomrule
\end{tabularx}

\caption{Ablation of the main performance drivers. For the autonomous-agent rows, the four single-component settings are variants built on top of the base tool, while ``Advanced tool'' combines all tool components.}
\label{tab:main_ablation}
\vspace{-0.4cm}
\end{table}

\subsubsection{Effect of Retrieval Depth}
Table~\ref{tab:retrieval_depth} shows the effect of retrieval depth on a random 700-question subset. Increasing depth from Top-10 to Top-30 yields only a marginal 0.8\% improvement. Balancing this limited gain against token budgets, we adopt Top-10 for the main experiments. Note that subset evaluation may cause slight deviations from our main results. We also ablate the embedding models in Appendix~\ref{tab:embedding_ab}.

\begin{table}[t]
\centering
\small
\begin{tabular}{lc}
\toprule
\textbf{Retrieval Depth} & \textbf{Hits@1} \\
\midrule
Top-10 & 0.898 \\
Top-15 & 0.898 \\
Top-20 & 0.904 \\
Top-25 & 0.904 \\
Top-30 & 0.906 \\
\bottomrule
\end{tabular}
\caption{Effect of retrieval depth.}
\label{tab:retrieval_depth}
\end{table}

\subsubsection{Backbone Generality}

Finally, we evaluate the zero-shot autonomous framework with different backbone LLMs. 
As shown in Table~\ref{tab:backbone}, it remains effective across all tested backbones, including Qwen3-Max (79.3\%), Kimi-2.5 (78.9\%), DeepSeek-R1 (84.2\%), and DeepSeek V3.2 (84.4\%). 
Even in the zero-shot setting, all variants surpass the previous state-of-the-art (78.0\%), indicating that the framework generalizes well across backbone models.
\begin{table}[t]
\centering
\small
\begin{tabular}{lc}
\toprule
\textbf{Backbone LLM (Zero-shot)} & \textbf{Hits@1} \\
\midrule
Qwen3-Max & 0.793 \\
DeepSeek-R1 & 0.842 \\
Kimi-2.5 & 0.789 \\
DeepSeek V3.2 & \textbf{0.844} \\
\bottomrule
\end{tabular}
\caption{Zero-shot performance across different LLMs.}
\label{tab:backbone}
\vspace{-0.4cm}
\end{table}


\subsection{Analysis of Interpretability}
\label{sec:interpretability-analysis}

AT2QA provides a complete evidence chain for its prediction process, including retrieval, filtering, fact selection, and answer determination, rather than exposing only the final answer. 
Figures~\ref{fig:equal-multi-case-study}, \ref{fig:before-last-case-study}, and \ref{fig:self-correction-case-study} present representative case studies, where the highlighted facts show the key supporting evidence. 
Beyond this, Appendix~\ref{subsec:behavioral_analysis} provides additional supporting evidence that autonomy brings extra benefits by inducing self-correction and self-validation, thereby fully unlocking the model's capabilities.

\section{Conclusion}
In this paper, we introduced \textbf{AT2QA}, an autonomy-first and training-free agentic framework for temporal knowledge graph question answering. Departing from fixed, human-designed workflows and expensive fine-tuning pipelines, AT2QA enables the LLM to determine \emph{which actions are necessary next} while iteratively interacting with the TKG environment through a generic search tool. This design inherently allows the agent to dynamicly verify retrieved evidence, reformulate queries, and self-correct its reasoning trajectory when errors or contradictions arise. To further elicit complex temporal reasoning without gradient updates, we proposed a training-free experience mining strategy that distills a compact few-shot library from successful self-generated trajectories. Experiments on three challenging benchmarks show that AT2QA achieves state-of-the-art results, while also producing transparent and verifiable reasoning traces.

\section*{Limitations}
\paragraph{Efficiency and scalability.}
Our implementation performs nearest-neighbor retrieval over a structurally filtered candidate set and allows up to $T_{\max}=20$ interaction turns. While this design improves robustness, it increases latency and inference cost compared to single-pass RAG. Scaling to substantially larger graphs or tighter latency budgets may require more efficient indexing (e.g., ANN) and better turn-level early stopping policies.

\paragraph{Autonomy can be inefficient and unstable.}
Granting the LLM full autonomy improves robustness, but it may also lead to \emph{extra exploration turns} or occasional looping behaviors under ambiguous queries or strong distractors. As a result, latency and inference cost can be higher than single-pass RAG or fixed pipelines, and performance may be more sensitive to decoding randomness and stopping criteria.

\bibliography{anthology,custom}

\appendix
\section{Dataset Details}
\label{sec:appendix-Dataset}
\paragraph{MultiTQ.}
Constructed from the ICEWS05-15 dataset, MultiTQ is a large-scale benchmark for multi-granularity temporal question answering. It contains approximately 500K question-answer pairs and more than 461K temporal facts represented as quadruples. The dataset spans multiple temporal granularities, including years, months, and days, and requires models to handle diverse temporal reasoning patterns under explicit or implicit constraints. Following the original benchmark design, questions are organized into six categories: \textit{Equal}, \textit{Before/After}, and \textit{First/Last} under the \textit{Single} setting, as well as \textit{Equal Multi}, \textit{After First}, and \textit{Before Last} under the \textit{Multiple} setting. These categories cover direct temporal retrieval, comparative reasoning, and multi-hop temporal inference over multiple entities. Owing to its large scale, diverse temporal granularity, and strong compositionality, MultiTQ provides a challenging testbed for evaluating AT2QA’s autonomous search, temporal grounding, and consistency checking abilities. The detailed statistics are summarized in Table~\ref{tab:multitq}.
\begin{table}[t]
\centering
\resizebox{\columnwidth}{!}{%
\begin{tabular}{llrrr}
\toprule
\multicolumn{2}{c}{\textbf{Category}} & \textbf{Train} & \textbf{Dev} & \textbf{Test} \\
\midrule
\multirow{3}{*}{Single} & Equal & 135,890 & 18,983 & 17,311 \\
 & Before/After & 75,340 & 11,655 & 11,073 \\
 & First/Last & 72,252 & 11,097 & 10,480 \\
\midrule
\multirow{3}{*}{Multiple} & Equal Multi & 16,893 & 3,213 & 3,207 \\
 & After First & 43,305 & 6,499 & 6,266 \\
 & Before Last & 43,107 & 6,532 & 6,247 \\
\midrule
\multicolumn{2}{c}{\textbf{Total}} & \textbf{386,787} & \textbf{57,979} & \textbf{54,584} \\
\bottomrule
\end{tabular}%
}
\caption{Statistics of question categories in MultiTQ.}
\label{tab:multitq}
\end{table}
\paragraph{Timeline-CronQuestion.}
Derived from the CronQuestion knowledge graph, Timeline-CronQuestion is a TimelineKGQA benchmark designed for temporal question answering over time-point-centric knowledge graphs. It contains 41,720 question-answer pairs. Compared with MultiTQ, this benchmark places stronger emphasis on timeline-centric temporal reasoning, especially temporal arithmetic and semantic operations over intervals. In particular, models must handle duration reasoning, interval composition, and set-like operations over temporal spans, with answers extending beyond entities and timestamps to include time ranges or durations. Following the TimelineKGQA taxonomy, questions are grouped into three difficulty levels: \textit{Simple}, \textit{Medium}, and \textit{Complex}, corresponding to reasoning over one, two, and multiple context events, respectively. This balanced difficulty structure makes Timeline-CronQuestion well suited for evaluating whether AT2QA can generalize from direct temporal retrieval to compositional temporal inference. The detailed statistics are summarized in Table~\ref{tab:timeline-cronquestion}.

\begin{table}[t]
\centering
\small
\begin{tabular*}{\columnwidth}{@{\extracolsep{\fill}}ccccc@{}}
\toprule
\multicolumn{2}{c}{\textbf{Category}} & \textbf{Train} & \textbf{Dev} & \textbf{Test} \\
\midrule
\multicolumn{2}{c}{Simple}  & 7,200  & 2,400 & 2,400 \\
\multicolumn{2}{c}{Medium}  & 8,252  & 2,751 & 2,751 \\
\multicolumn{2}{c}{Complex} & 9,580  & 3,193 & 3,193 \\
\midrule
\multicolumn{2}{c}{\textbf{Total}} & \textbf{25,032} & \textbf{8,344} & \textbf{8,344} \\
\bottomrule
\end{tabular*}
\caption{Statistics of question categories in Timeline-CronQuestion.}
\label{tab:timeline-cronquestion}
\vspace{-0.2cm}
\end{table}

\paragraph{Timeline-ICEWS-Actor.}
Constructed from the ICEWS Coded Event Data, Timeline-ICEWS-Actor is a TimelineKGQA benchmark for actor-centric temporal question answering over dynamic event sequences. It contains 89,372 question-answer pairs. In contrast to Timeline-CronQuestion, this dataset is grounded in event-centric political interactions and focuses more directly on reasoning about actors, event positions, and temporally evolving relations in a dynamic timeline. Questions are likewise divided into three difficulty levels---\textit{Simple}, \textit{Medium}, and \textit{Complex}---which require progressively more challenging temporal reasoning over one, two, and multiple events. Its relatively balanced distribution across difficulty levels, together with its event-centric structure, makes Timeline-ICEWS-Actor a valuable benchmark for assessing AT2QA’s robustness in multi-step timeline reasoning and actor-focused evidence aggregation. The detailed statistics are summarized in Table~\ref{tab:timeline-icews-actor}.
\begin{table}[t]
\centering
\small
\begin{tabular*}{\columnwidth}{@{\extracolsep{\fill}}ccccc@{}}
\toprule
\multicolumn{2}{c}{\textbf{Category}} & \textbf{Train} & \textbf{Dev} & \textbf{Test} \\
\midrule
\multicolumn{2}{c}{Simple}  & 17,982 & 5,994 & 5,994 \\
\multicolumn{2}{c}{Medium}  & 15,990 & 5,330 & 5,330 \\
\multicolumn{2}{c}{Complex} & 19,652 & 6,550 & 6,550 \\
\midrule
\multicolumn{2}{c}{\textbf{Total}} & \textbf{53,624} & \textbf{17,874} & \textbf{17,874} \\
\bottomrule
\end{tabular*}
\caption{Statistics of question categories in Timeline-ICEWS-Actor.}
\label{tab:timeline-icews-actor}
\vspace{-0.4cm}
\end{table}

\section{Quantitative Analysis of Agentic Behaviors}
\label{subsec:behavioral_analysis}
A fundamental limitation of static LLM workflows is their inability to recover from intermediate retrieval errors. To prove that our massive performance gain on multi-hop questions stems from genuine agentic behaviors—specifically \textit{Self-Correction} and \textit{Self-Validation}—we conduct a quantitative trajectory analysis.

\begin{figure}[htbp]
    \centering

    \includegraphics[width=1.0\linewidth]{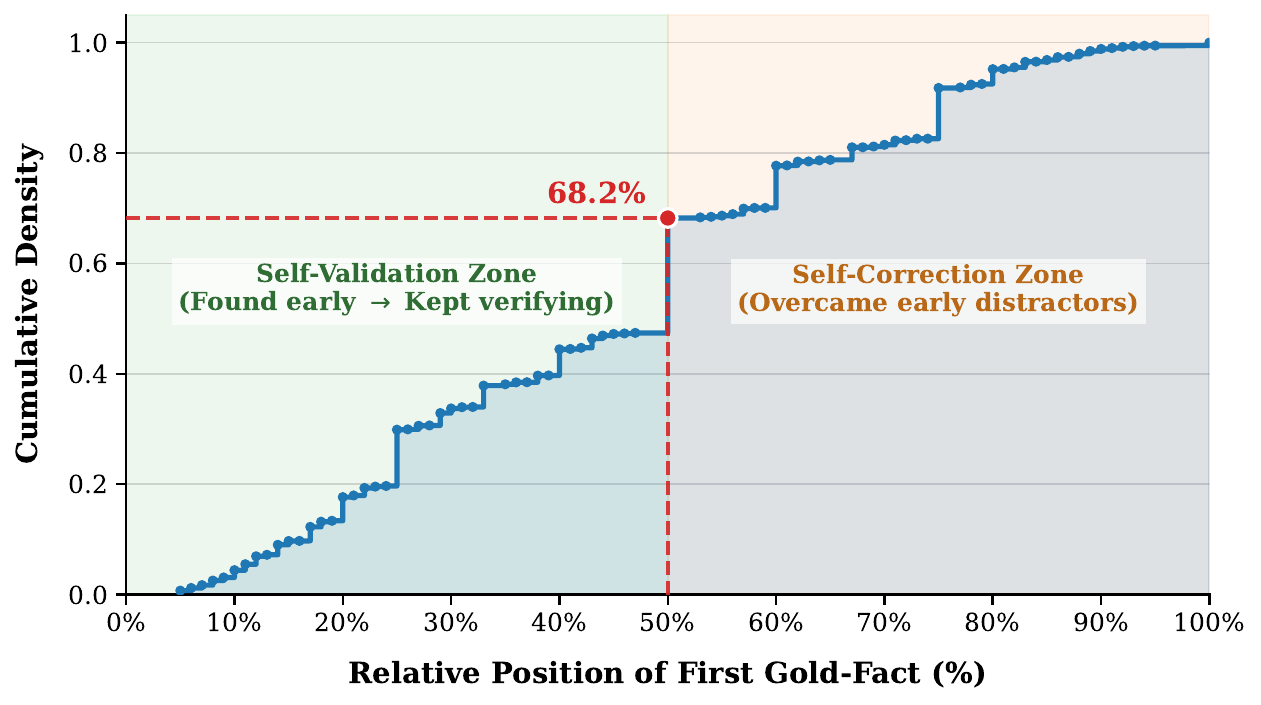}
    \caption{Cumulative Density of First Gold-Fact Position for complex multiple-target queries. The distribution provides quantitative proof of the agent's self-validation (left) and self-correction (right) capabilities.}
    \label{fig:cdf_trajectory}
\end{figure}

We define a \textit{Gold Fact} as the specific retrieved evidence required to deduce the correct answer. We track the relative position (in percentage) of the \textbf{first appearance of the Gold Fact} within the agent's entire interaction trajectory. Figure~\ref{fig:cdf_trajectory} plots the Cumulative Density Function (CDF) of this metric for all successfully answered \textit{Multiple-target} questions that involved more than 3 reasoning rounds.

The CDF curve reveals two distinct and profound agentic behaviors:
\paragraph{Self-Correction (The Right Tail).} 
Remarkably, in approximately 32\% of the successful cases, the very first Gold Fact does not appear until the \textit{latter half} (>50\%) of the total reasoning rounds. This indicates that during the early stages of navigation, the agent frequently retrieved irrelevant, distracting, or contradictory facts. Instead of hallucinating an answer based on this noisy context, the agent autonomously recognized the failure, adjusted its search constraints (e.g., modifying temporal windows or entity roles), and repeatedly tried until the correct evidence was surfaced. This confirms a robust \textbf{Self-Correction} mechanism.
\paragraph{Self-Validation (The Left Tail).} 
Conversely, in cases where the Gold Fact was discovered early in the trajectory (<50\%), the agent \textit{did not} immediately terminate the session. Since the questions demand multiple answers, premature termination would lead to partial failures. The log shows that the agent retained the initial Gold Fact in its memory, recognized that the evidence was insufficient to holistically answer the query, and deliberately continued searching to verify and gather the remaining pieces. This \textbf{Self-Validation} behavior demonstrates a high degree of meta-cognitive planning, sharply contrasting with naive RAG pipelines that stop after a single semantic match.

\subsection{Impact of Backbone Embedding Models}
\begin{table}[t]
\centering
\small
\begin{tabular}{lc}
\toprule
\textbf{Backbone Embedding Model (Zero-shot)} & \textbf{Hits@1} \\
\midrule
text-embedding-3-large & 0.883 \\
text-embedding-3-small & 0.892 \\
gemini-embedding-001 & 0.889 \\
Baidu-Embedding-V1 & 0.877 \\
\bottomrule
\end{tabular}
\caption{Performance across different embedding.}
\label{tab:embedding_ab}
\vspace{-0.4cm}
\end{table}

To investigate whether our framework's performance heavily relies on the semantic matching capabilities of a specific dense retriever, we conducted an ablation study evaluating the impact of different backbone embedding models. To ensure a fair and efficient comparison, this evaluation was performed on a fixed subset of 500 questions randomly sampled from the test set.

We substituted the default embedding model with several leading alternatives, including OpenAI's text-embedding-3-large and text-embedding-3-small, Google's gemini-embedding-001, and Baidu-Embedding-V1. All other configurations, including the LLM agent and the maximum tool interaction rounds, remained identical.

As shown in Table \ref{tab:embedding_ab}, the choice of embedding model has a negligible impact on the final performance. The Hits@1 scores remain highly stable across all tested models, fluctuating tightly within a narrow margin of 1.5\% (from 0.877 to 0.892). Notably, lighter models such as text-embedding-3-small perform on par with, or even slightly surpass, heavier models. This suggests that the minor 1.5\% variance is primarily attributable to random fluctuations

\section{Additional Analysis on Single-Gold Undercounting in Timeline-CronQuestion}
\label{sec:appendix-TimelineCron}
During error analysis, we found that a substantial portion of AT2QA's officially counted errors on Timeline-CronQuestion are factually correct predictions that are penalized by the benchmark's single-gold exact-match evaluation.
In particular, some questions admit multiple valid answers in the underlying temporal knowledge graph, while the dataset keeps only one of them as the annotated gold answer.
As a result, a prediction can be correct with respect to the question and graph facts, yet still be counted as incorrect if it does not exactly match the single annotated answer.

We manually audited AT2QA's officially incorrect predictions on the test set.
For the \textbf{Simple} split, 371 out of 406 official errors (91.4\%) were found to be factually correct answers excluded by the single-gold annotation.
For the \textbf{Medium} split, the same issue was observed in 226 out of 1,012 official errors (23.0\%).
We also observed a smaller subset of cases in which the annotated gold answer itself appears to be incorrect.
Representative examples are shown in Table~\ref{tab:timeline_cronquestion_case_studies}.

A likely reason is the question construction process of Timeline-CronQuestion: questions are generated from a fixed subset of sampled temporal facts, while other graph facts that also satisfy the same constraints may be omitted from the final answer annotation.
This can lead to under-specified answer sets and consequently undercount factual correctness under exact-match evaluation.

These findings suggest that the official exact-match score may underestimate AT2QA's factual correctness on Timeline-CronQuestion, especially on the Simple and Medium splits.
Nevertheless, all main results in this paper are reported strictly under the official benchmark protocol; this analysis is intended only as a diagnostic supplement rather than a replacement for the standard evaluation.
\begin{table}[t]
\centering
\scriptsize
\setlength{\tabcolsep}{4pt}
\renewcommand{\arraystretch}{1.15}
\begin{tabularx}{\columnwidth}{@{}p{0.11\columnwidth} p{0.16\columnwidth} Y Y@{}}
\toprule
\textbf{QLevel} & \textbf{Reason} & \textbf{Example} & \textbf{AT2QA Answer Relevant Facts} \\
\midrule

\textbf{Simple} 
& insufficient gold (91.4\%) 
& \textbf{Question:} ``Burnley F.C. is member of sports team by who from 1960-01-01 to 1968-01-01?''

\textbf{Gold:} Willie Morgan

\textbf{AT2QA:} Andy Lochhead
& \textbf{Search fact:} 301957\fsep Andy Lochhead\fsep member of sports team\fsep Burnley F.C.\fsep 1960-01-01\fsep 1968-01-01
\\

\midrule

\textbf{Medium} 
& insufficient gold (23\%) 
& \textbf{Question:} ``Cell 211 nominated for which object after Romeo Menti member of sports team A.C. Milan?''

\textbf{Gold:} Goya Award for Best Film

\textbf{AT2QA:} Goya Award for Best Producer
& \textbf{Search fact:} 120994\fsep Cell 211\fsep nominated for\fsep Goya Award for Best Producer\fsep 2010-01-01\fsep 2010-01-01

289197\fsep Romeo Menti\fsep member of sports team\fsep A.C. Milan\fsep 1944-01-01\fsep 1944-01-01
\\

\bottomrule
\end{tabularx}
\caption{Representative examples of answer undercounting in Timeline-CronQuestion.}
\label{tab:timeline_cronquestion_case_studies}
\vspace{-0.4cm}
\end{table}






\section{Case Study}
\label{sec:appendix}
To provide a deeper understanding of how AT2QA autonomously navigates TKGs to execute complex temporal reasoning, we present a qualitative analysis of several representative cases. These studies intuitively demonstrate the agent's robust dynamic self-correction and strict consistency checking capabilities, all achieved without any parameter updates. 

\begin{figure*}[t]
  \centering
  \includegraphics[
    page=1,
    width=\textwidth,
    trim=49 35 52 51,
    clip
  ]{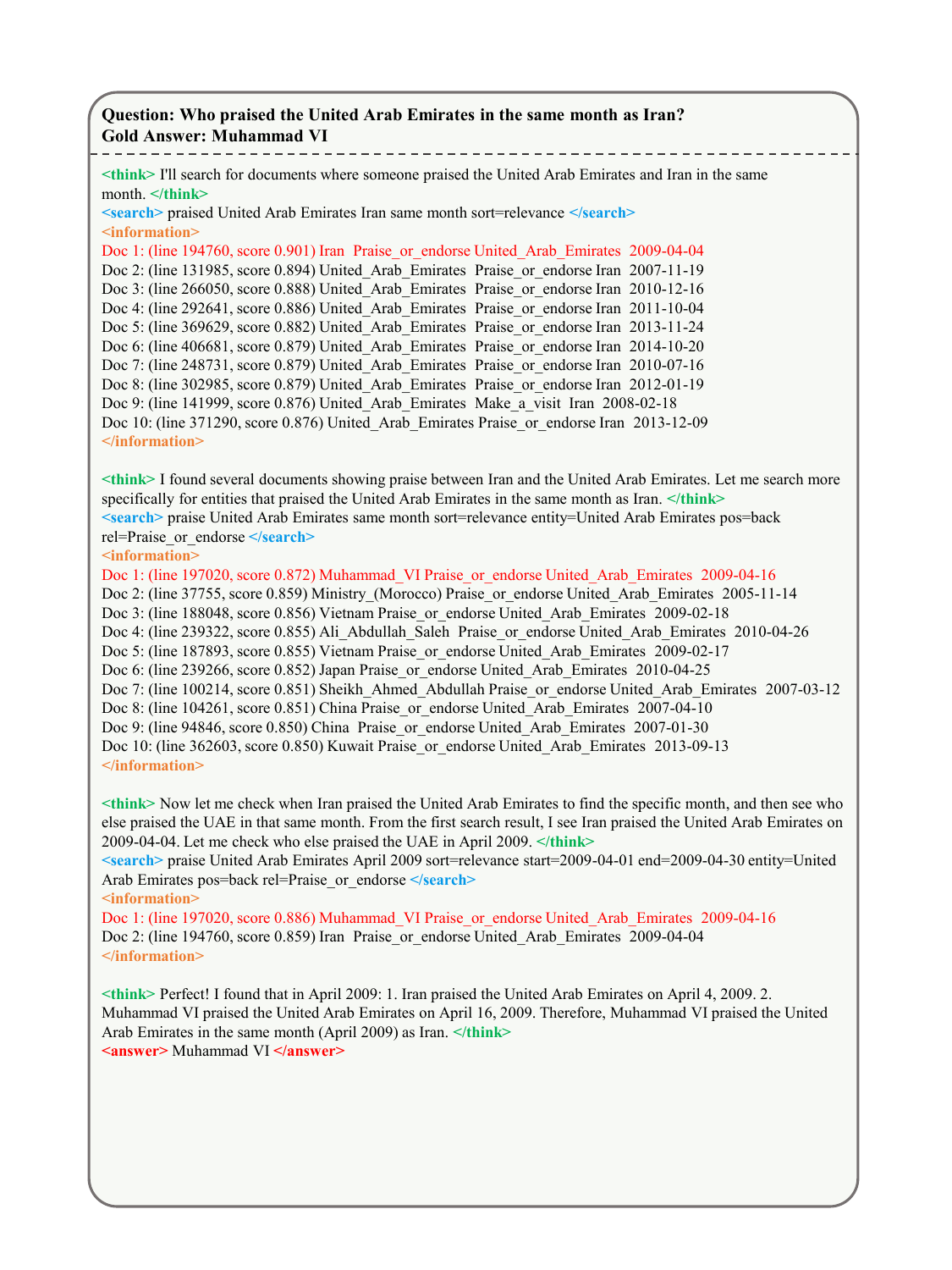}
  \caption{Case Study for Equal Multi Questions.}
  \label{fig:equal-multi-case-study}
\end{figure*}

\begin{figure*}[t]
  \centering
  \includegraphics[
    page=1,
    width=\textwidth,
    trim=50 35 51 51,
    clip
  ]{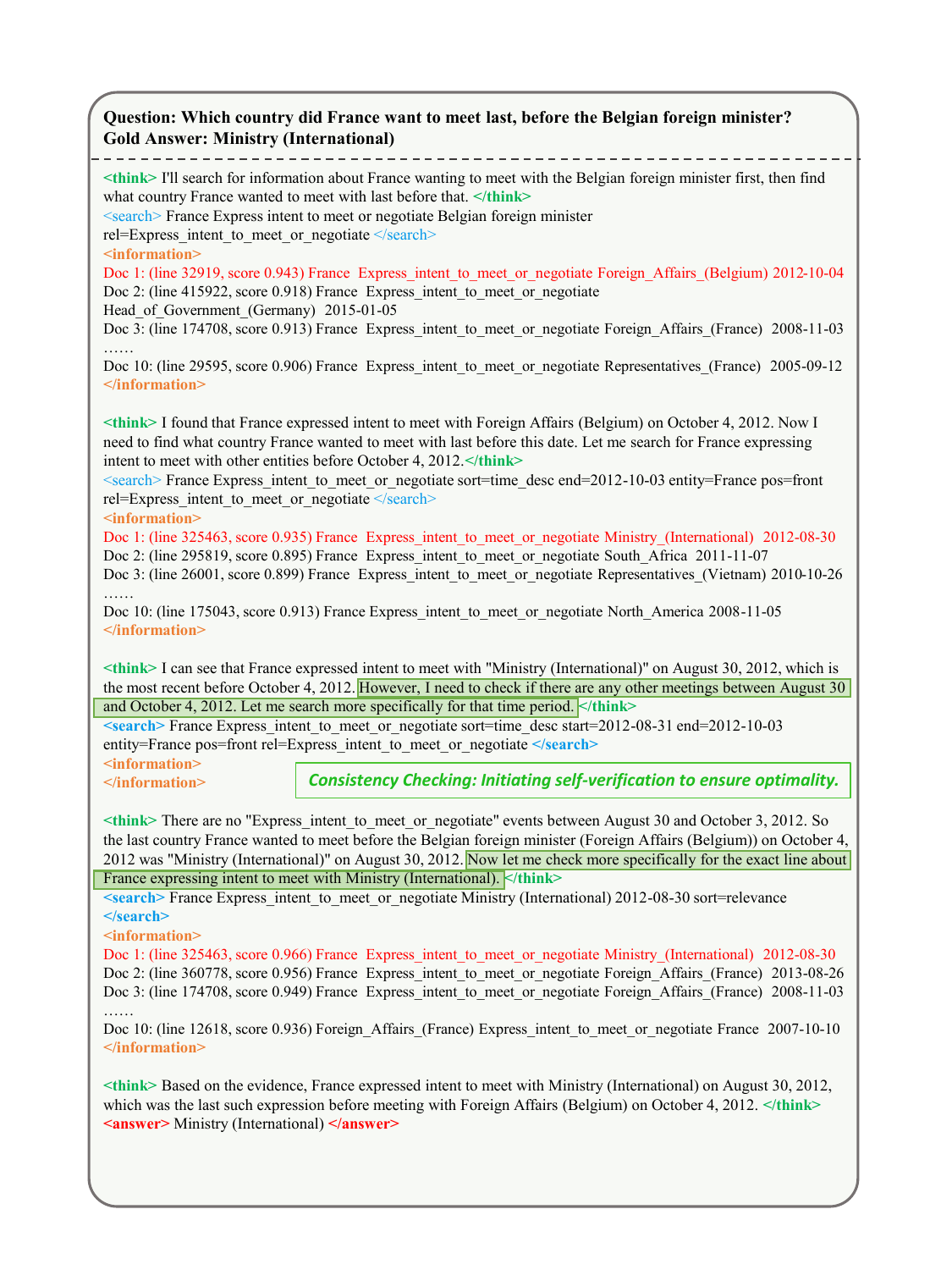}
  \caption{Case Study for Before Last Questions with Consistency Checking.}
  \label{fig:before-last-case-study}
\end{figure*}

\begin{figure*}[t]
  \centering
  \includegraphics[
    page=1,
    width=\textwidth,
    trim=49 35 52 50,
    clip
  ]{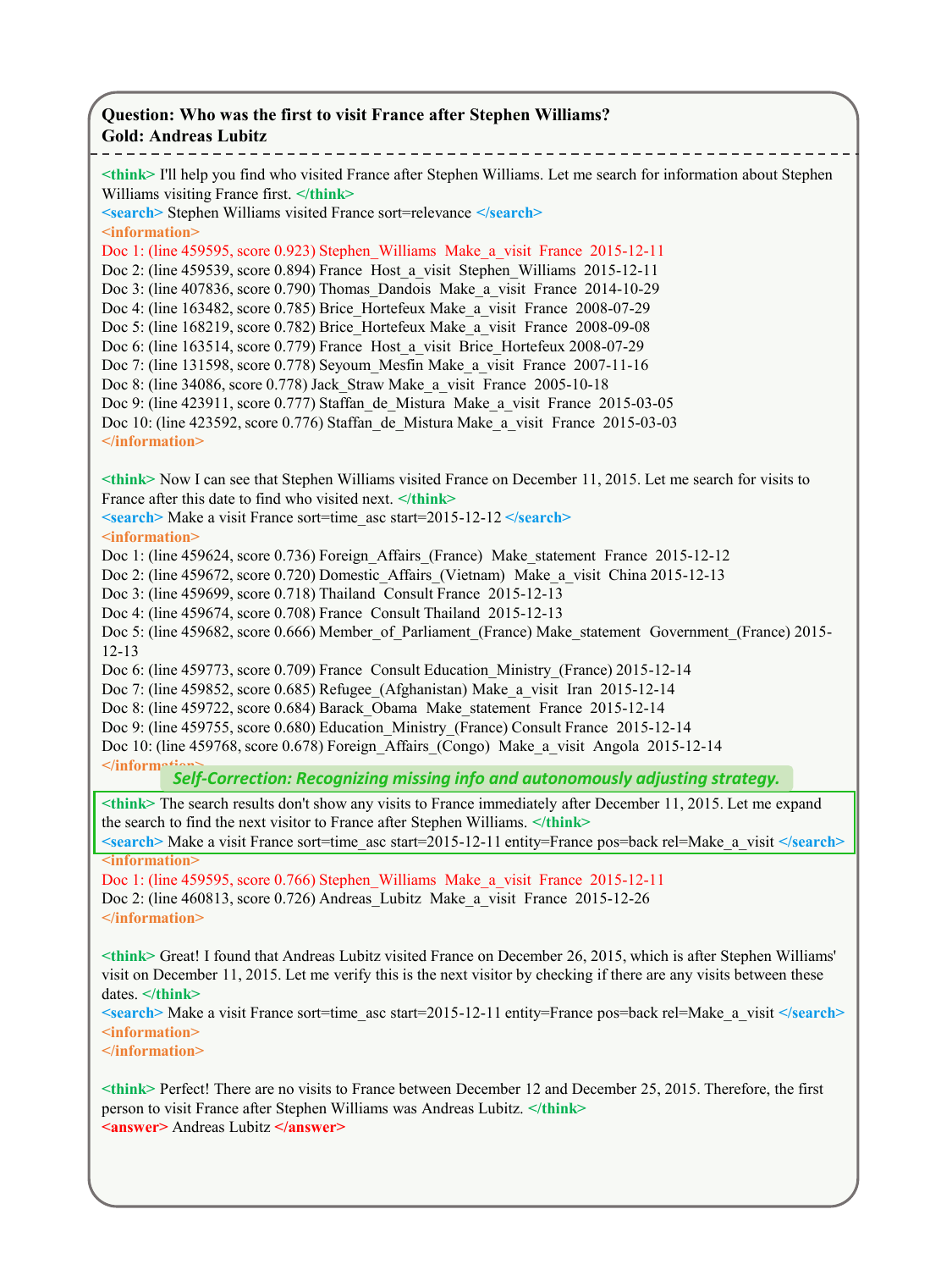}
  \caption{Case Study for After First Questions with Self-Correction.}
  \label{fig:self-correction-case-study}
\end{figure*}

Figure~\ref{fig:equal-multi-case-study} illustrates the reasoning trajectory for an "Equal Multi" question from the MultiTQ dataset. This example highlights AT2QA's proficiency in accurately parsing implicit temporal constraints and leveraging the time window of a pivot event to effectively bound the search space for concurrent multi-entity activities.
Figures~\ref{fig:before-last-case-study} and~\ref{fig:self-correction-case-study} detail the reasoning chains for the highly challenging "Before Last" and "After First" question types, respectively. In these scenarios, AT2QA explicitly grounds the anchor event first, and subsequently conducts fine-grained evaluations of subsequent events within strictly defined temporal boundaries.
Figure~\ref{fig:before-last-case-study} illustrates AT2QA's consistency checking mechanism. Upon retrieving a highly plausible candidate answer (i.e., "Ministry (International)"), the agent resists a greedy acceptance strategy. Instead, it proactively initiates a more granular internal verification over the specified time slice to guarantee that no subsequent entities visited France within the window, thereby ensuring the global optimality of the final conclusion.
Furthermore, Figure~\ref{fig:self-correction-case-study} demonstrates dynamic self-correction capability of AT2QA. Upon detecting that the current retrieval strategy fails to recall the requisite evidence, the agent autonomously rolls back its reasoning state and dynamically rewrites its query strategy until the crucial evidence is successfully pinpointed. This closed-loop correction mechanism effectively breaks the bottleneck of cascading error amplification inherent in traditional rigid workflows.

\end{document}